**RESEARCH**                                                                 **Open Access**

# Supervised multiple kernel learning approaches for multi-omics data integration

Mitja Briscik[1*†], Gabriele Tazza[2*†], László Vidács[2], Marie-Agnès Dillies[3] and Sébastien Déjean[1]

†Mitja Briscik and Gabriele Tazza contributed equally to this work.

*Correspondence:
mitja.briscik@math.univ-toulouse.fr; tazza@inf.u-szeged.hu

[1] Institut de Mathématiques de Toulouse, UMR5219, CNRS, UPS, Université de Toulouse, Cedex 9, Toulouse 31062, France
[2] Department of Computer Science, Applied Artificial Intelligence Group, University of Szeged, Szeged 6720, Hungary
[3] Institut Pasteur, Université Paris Cité, Bioinformatics and Biostatistics Hub, F-75015 Paris, France

## Abstract

**Background:** Advances in high-throughput technologies have originated an ever-increasing availability of omics datasets. The integration of multiple heterogeneous data sources is currently an issue for biology and bioinformatics. Multiple kernel learning (MKL) has shown to be a flexible and valid approach to consider the diverse nature of multi-omics inputs, despite being an underused tool in genomic data mining.

**Results:** We provide novel MKL approaches based on different kernel fusion strategies. To learn from the meta-kernel of input kernels, we adapted unsupervised integration algorithms for supervised tasks with support vector machines. We also tested deep learning architectures for kernel fusion and classification. The results show that MKL-based models can outperform more complex, state-of-the-art, supervised multi-omics integrative approaches.

**Conclusion:** Multiple kernel learning offers a natural framework for predictive models in multi-omics data. It proved to provide a fast and reliable solution that can compete with and outperform more complex architectures. Our results offer a direction for bio-data mining research, biomarker discovery and further development of methods for heterogeneous data integration.

## Introduction

Data integration has recently attracted substantial attention in the research literature, both for the statistical challenges and promising potential applications in fields such as biology and medicine. Multi-omics data have become increasingly available following the significant growth of high-throughput technologies. The availability of such rich while complex data has expanded the number of available algorithms and methodologies to properly conduct analyses, with the possible need to create novel research profiles [1]. In this context, Kernel methods have proven to be a very promising technique for integrating and analyzing high-throughput technologies-generated data. Kernel methods benefit from the possibility of providing a nonlinear version of any linear algorithm that relies solely on dot products. For instance, unsupervised methods such as Kernel Principal Component Analysis [2], Kernel Canonical

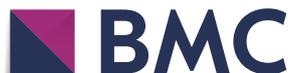

Correlation Analysis [3], Kernel Discriminant Analysis [4] and Kernel Clustering [5] are all examples of nonlinear algorithms enabled by the so-called kernel trick.

Kernel-based methods also include supervised classification algorithms. Support vector machine is the most popular one, along with Kernel partial least squared regression [6] or Kernel discriminant analysis [7].

Several methodologies are also available to integrate multiple high throughput data sources through the so-called Multiple kernel learning (MKL) approach. These methods combine modern optimization techniques' power with kernel methods' framework, providing a new multi-source genomic data learning tool.

In this work, we review classical MKL algorithms, while also exploring alternative MKL approaches. Specifically, we propose a novel approach that consists of adapting unsupervised algorithms for multiple kernel integration to a supervised context, i.e., fitting an SVM classification model on a fused kernel obtained through an unsupervised algorithm for the convex linear combination of input kernels. This approach mimics what more recent deep learning-based methods realize using Autoencoders [8]. First, the lower dimensional latent representation is learned in an unsupervised way by an Autoencoder, and then this embedding is used to perform a downstream task such as classification [9].

More recently, Deep learning has emerged as a valid alternative to dealing with data integration challenges. A key strength of deep learning lies in its ability to learn homogeneous representations from heterogeneous data sources (images, text, tabular data), making it a perfect candidate for multi-omics integration problems.

Different deep learning methods have already been applied in this domain with promising results. Architectures such as Autoencoders [8, 10], Graph Neural Networks [11, 12] or Multi-head Attention [13] have been successfully adapted to different multi-omics integration tasks reaching the state-of-the-art.

Deep learning has also been used as an alternative approach to multiple kernel fusion [14] to integrate different kernels from a single data source. This type of architecture can be easily adapted to integrate heterogeneous data sources, such as multi-omics datasets. With this intention, we introduce a novel deep learning framework tailored for Multiple kernel learning (MKL), namely DeepMKL, specifically within multi-omics integration. This method exploits both the advantages of kernel learning and deep learning by transforming the input omics using different kernel functions and guiding their integration in a supervised way, optimizing the neural network weights to minimize the classification error.

To sum up, while Multiple kernel learning remains an under-utilized tool for genomic data mining [15], in this work, we propose MKL methods to integrate multi-omics data based both on unsupervised convex linear optimization and deep learning. We aim to show the advantages of this setting by comparing it with state-of-the-art methods. Our results align with recent findings in Brouard et al. [16], where the authors compare traditional machine learning (ML) models with Graph Neural Networks (GNNs) in single omics analysis, concluding that the benefits of GNNs are overstated. We similarly demonstrate that classical ML approaches, such as MKL methods, show competitive results against GNNs in the context of multi-omics analysis.

## Related work

Many machine learning methods are available to unravel biological system mechanisms and find new biomarkers. The big challenges associated with multi-omics data mining and integration are the intrinsic high dimensionality, heterogeneity and nonlinearity of the sample space. For this reason, refined methods are needed to give practitioners new direction and solutions for analyzing such complex datasets.

Numerous integration strategies are available in the literature, including early, mixed, hierarchical, intermediate and late integration. In this work, we focus on the mixed integration type, which has demonstrated to ensure great adaptability for omics data fusion as reviewed in Picard et al. [17].

Early stage integration, the easiest and fastest procedure available, nonetheless poses intrinsic drawbacks. More specifically, since early integration is based on the concatenation of the original data, it naturally increases the input dimensionality while giving more importance to omics with a bigger number of features. Moreover, while being extremely easy and fast to realize, this practice tends to mislead learning algorithms as it does not consider the specific data distribution of each input dataset.

On the contrary, mixed integration allows ML algorithms to conduct the learning phase on more refined and less dimensional datasets. As these methods produce new versions of the input datasets which are more homogeneous than original versions, it facilitates ML algorithms to operate on a unified single input for learning.

Furthermore, another very popular strategy is late integration, which consists of applying each machine learning model separately on each input dataset and then of combining their respective predictions in a later stage. However, as claimed in Picard et al. [17], this approach may not be relevant for biological applications. Indeed, an integration based solely on the combination of different model predictions cannot be compared to a procedure that directly considers complementary information among different omics, as it can be seen as a multiple single-omics analyses.

In the present work, we will investigate mixed integration techniques for multi-omics data integration in comparison to the state-of-the-art method i.e. MOGONET in Wang et al. [12], a late integration methodology based on GNNs.

### Mixed integration

It is generally accepted that a classification model trained with information obtained from different sources leads to a more comprehensive overview of the problem [18, 19].

In the field of omics sciences, when different data obtained on the same individuals are available, the integrated analysis can provide richer information about the biological system compared to the results achieved using a single layer of information. New achievements have been reached in a wide area of research, for instance allowing the identification of molecular signatures of human breast tumours [20] or for microbial communities profiling [21].

Each omic dataset contains a different aspect of the mechanisms regulating a biological phenotype. In addition, the technologies used to collect them differ. Consequently, the nature and structure of those data are usually very diverse, generating a remarkably heterogeneous framework. Mixed integration or transformation-based strategies undertake the flaws of concatenation-based approaches applying ML algorithm to a simpler representation of each input dataset. The original omics are transformed separately to obtain a clearer, richer and lower in dimensions version. Standard transformation methods that can be used are kernel-based, graph-based, and deep learning methods.

In this work we will focus our attention on kernel-based integration and on deep learning-based methods applied on kernel learning.

*Multiple kernel learning*

Kernel methods have been shown to offer an elegant and natural mathematical solution to address data integration from heterogeneous sources, as using kernels enables the representation of the datasets in terms of pairwise similarities between sample points [22, 23]. Given a dataset of *n* observations $x_1, \ldots, x_n$ with $x_i \in \mathbb{R}^p$, a function k defined as k: $\mathbb{R}^p \times \mathbb{R}^p \longrightarrow \mathbb{R}$ is a valid kernel if it is symmetric and positive semi-definite i.e. $k(x_i, x_j) = k(x_j, x_i)$ and $c^T K c \geqslant 0$, $\forall c \in \mathbb{R}^n$, where $K$ is the $n \times n$ kernel matrix containing all the data pairwise similarities $K = k(x_i, x_j)$.

Every kernel function is associated with an implicit function $\phi$: $\mathbb{R}^p \longrightarrow \mathcal{H}$ which maps the input points into a generic feature space $\mathcal{H}$, with possibly an infinite dimensionality, with the expression $k(x_i, x_j) = \langle \phi(x_i), \phi(x_j) \rangle$. This relation allows the implicitly computation of the dot products in the feature space by applying the kernel function to the input objects, without explicitly computing the mapping function $\phi$ [24].

It is generally accepted that the sample space of many research problems, such as omics data, is often nonlinear [25]. This nonlinearity is linked also to the incomplete understanding, for instance, of gene interactions and biological pathways, which suggests that genes are not connected in a simple linear way. In this context, kernel methods offer a natural and not computationally expensive approach to kernelized i.e. obtain nonlinear version of any algorithm purely based on dot-product calculations. Indeed, by replacing the linear dot product in the input space by the kernel pairwise values, it is possible to implicitly obtain the value of the dot product as it was computed directly in the feature space. This is the so-called *kernel trick*, which allows algorithms designed initially for linear data to be extended to nonlinear frameworks by implicitly mapping the input points into high-dimensional feature spaces induced by the kernel.

In the context of multi omics integration, given different datasets based on the same *n* observations, kernel methods provide another advantage, namely they allow to represent every original dataset with a $n \times n$ kernel matrix $K$. So, even if the original data types are heterogeneous (counts, factors, continuous data, networks, images), after the kernel transformation, all the *M* input datasets will have the form of a $n \times n$ matrix with real numbers as entries, with *M* equal to the number of available omic datasets.

Moreover, the meta-kernel obtained from the combination of the *M* input kernels is a global similarity matrix containing the sample's similarities based on the original datasets' variables. MKL assures great adaptability as many kernel functions are available, such as linear, Gaussian, polynomial, or sigmoid. In this way it is possible to choose and to apply a specific kernel function on a certain omic input, as each function may be more suitable for a specific omic.

The most common approach in Multiple kernel learning is to compute a convex linear combination of kernel Gram matrices. Analytically, given *M* different datasets, MKL consists of the linear combination of the *M* kernel matrices, as in

$$K^* = \sum_{m=1}^{M} \beta_m K^m, \tag{1}$$

with $\beta_m \neq 0$ and $\sum_{m=1}^{M} \beta_m = 1$.

It directly follows that the simplest solution is to fix all the weights to be equal, i.e. to $\frac{1}{M}$. Of course, this setting does not allow us to benefit from the adaptability of the multiple kernel framework. All kernels will contribute equally to the classifier, not taking into account possible redundant or less informative sources of information. The experiment section will denote this setting as **MKL-naive**.

Contrarily, the $\beta_m$ weights can be optimized more appropriately. Usually, in supervised learning, they are tuned, minimizing the prediction error. The literature offers many algorithms for supervised MKL optimization. For instance, in the work in Lanckriet et al. [26], the weights are optimized with semidefinite programming techniques. The fused kernel is then used to train an SVM classifier, giving better performances than single omic analysis.

Another approach can be found in Rakotomamonjy et al. [27] where the convex linear combination is obtained through a weighted 2-norm regularization constrained formulation to promote a sparse kernel combination and using a subgradient descent for weights optimization. The so-called **SimpleMKL** method is available in the R package *RMKL* developed by Wilson et al. [15].

The *RMKL* package proposes several other algorithms such as **SEMKL**, Simple and Efficient MKL by Xu et al. [28] where the weights computation is based on the equivalence between group-lasso and MKL. Both SimpleMKL and SEMKL belong to the class of algorithms known as wrapper methods for Multiple kernel learning, thus updating kernel weights after each iteration.

A more sophisticated version of these wrapper methods specialized in the reduction of the number of SVM computations is **SpicyMKL** in Suzuki and Tomioka [29], which is a proximal minimization method that converges super-linearly. This algorithm is also implemented in the *RMKL* package under the name of DALMKL.

A different way to find the kernel coefficients in the convex linear combination of kernels can be found in Yang et al. [30] with **GA-fKPLS**, where the authors propose to compute the kernel parameters and weights using genetic algorithms.

A different approach to MKL is presented in Gönen and Alpaydin [31] and Gönen and Alpaydın [32], where the authors question the practice of assigning the same weight to a kernel over the whole input space. In this work, they propose a localized Multiple kernel learning **LMKL** based on the local selection of the appropriate kernel function, allowing to reduce the number of support vectors.

To be noted that these wrappers methods have been recently tested in Wilson et al. [15], where it has been shown that all these algorithms seem to have similar performance in the case of an analysis with few kernels.

Multiple kernel learning can also be used in the unsupervised learning framework. In this context, selecting appropriate criteria for weight optimization is less straightforward, as it cannot be based on a target variable of interest. In other words, as it is natural to optimize the weights through the minimization of the prediction error for supervised learning, the same does not apply in an unsupervised context. Hence, the algorithms available to effectively determine a strategy to guide the fusion process of the input kernels in an unsupervised framework are less numerous than in the supervised literature in Mariette and Villa-Vialaneix [33], the authors proposed **STATIS-UMKL**, a methodology to provide an approach to reach a consensus kernel based on the resemblance of the different kernels. Specifically, the meta kernel is defined by maximizing the average similarity between kernels, measured using their cosines according to the Frobenius dot product. The similarity matrix between two kernels $C = (C_{mm'})_{m,m'=1,...,M}$ gives insight into how the different kernels relate to each other, revealing whether they complement or provide distinct information. This matrix can then be used to derive the meta kernel $K^*$, which maximizes the overall similarity with all other kernels in the set.

We have previously introduced how kernels enable us to map data into a higher-dimensional feature space without explicitly computing that space. In this new space, data that are not linearly separable in the original input space may become linearly separable, making it easier to apply linear classification techniques. While kernel methods offer this advantage of making previously nonlinearly separable data linearly separable, this benefit comes with a trade-off. The original features are no longer explicitly accessible after the kernel transformation, as the data is represented through similarities in a new feature space. Consequently, this makes interpreting the model more challenging, as it becomes difficult to directly trace back the role of individual features in the transformed space to the original input variables without referring to a label. In this context in Mariette and Villa-Vialaneix [33], the authors proposed a method based on kernel PCA and random permutation to evaluate the importance of the original variables. Specifically the idea consists in recomputing the $K^m$ kernels after the permutation of all the values of the samples for a given measure *j*, obtaining a new kernel $\tilde{K}^{m,j}$. The *Crone-Corsby* distances of kernel matrices are then computed to assess which variables lead to the most significant differences between the original kernel and the new kernels $\tilde{K}^{m,j}$. Also, in Briscik et al. [34], the authors proposed *KPCA-IG*, an approach which provides a data-driven feature importance, where the influence of each original variable can be computed in the space of the kernel principal components as in the standard PCA. This method offers a computationally fast feature ranking methodology to identify the most relevant original variables, solely based on partial derivative of the kernel function.

### *Deep learning approaches*

Deep learning techniques are increasingly being employed in the context of multi-omics data analysis. One of the advantage of deep learning is its capacity to learn homogeneous representations from different input sources. In particular, multi-modal architectures allow the use of heterogeneous datasets, such as images, tabular data, time series, or graphs, to learn the underlying complex relationships among different aspects of a biological phenotype.

As reviewed in Stahlschmidt et al. [35], this kind of architecture is gaining popularity in the biomedical field, where data are becoming increasingly multi-modal. Recently, in this context, different works introduced approaches based on multi-modal deep learning to deal with different types of omics data, these multi-modal architectures are suited for both Mixed and Late integration strategies. As introduced, we will concentrate on Mixed integration approaches compared to the Late integration methods that can be regarded as the state-of-the-art for the datasets of interest in our analysis [12, 13, 36].

One of the most commonly used deep learning methods for Mixed integration strategies is Autoencoder. Autoencoder is an unsupervised deep learning method used to learn a latent representation of the data by minimizing the reconstruction error between the input and the reconstructed output. In the context of Mixed integration, they can be easily used to learn independent homogeneous latent representations to integrate them in a final shared layer [9]. Autoencoders can also be used to learn latent representations that depend on different omics inputs, as in Wu and Fang [8]. In this case, the approach uses Autoencoders in two different steps, first as a pre-processing for the two different inputs and then as an integration step, part of the learning process.

Other possible approaches for Mixed integration involve the use of feedforward neural networks. In particular, in Lin et al. [37], the authors built an architecture based on different encoding sub-networks to learn homogeneous representations from the different types of omics data, then a fusion step to create a concatenated representation of multi-omics, and finally, a classification sub-network is used to perform the cancer subtype classification. Alternatively, in Sharifi-Noghabi et al. [38], a similar architecture equipped with a triplet loss is used for drug response prediction.

Despite this, several state-of-the-art methods belong to the Late integration family, such as MOGONET by Wang et al. [12], MOADLN by Gong et al. [13] and Dynamics in Han et al. [36]. MOGONET transforms the input data into matrices of similarity among observations to build a graph structure and apply a Graph Convolutional Neural Network to each omic to obtain an initial prediction. After this first step, a View Correlation Discovery Network (VCDN) finally combines all the independent predictions to determine the correct label.

MOADLN, instead, uses the Self-attention mechanism to build a similarity network and exploit the correlation between intra-omic observations. In this case, each input instance is an element of a set i.e. a specific observation within a single omics type, and the Self Attention mechanism learns the weights for each of these elements, meaning that it determines the significance of each instance in relation with others within the same omics type. Also, for MOADLN, the first step is the initial independent prediction for each omics type, followed by a final combination through a Multi-Omics Correlation Discovery Network (MOCDN) to explore the cross-omics relations. Dynamics assesses feature-level and modality-level informativeness dynamically across different samples. It incorporates a sparse gating mechanism to capture variations in features information within each omics while using actual class probability to asses the classification confidence at the modality level [36].

### Materials and methods

As considered in the previous section, Wang et al. [12] and Gong et al. [13] claim to be the state-of-the-art in terms of predictive performance.

In this section, we present all the experiments to test different MKL methods, architectures and combinations in order to compare possible solutions for multi-omics data integration.

## Datasets

The datasets considered in this work are the publicly available ROSMAP for Alzheimer's Disease classification, BRCA for breast invasive carcinoma PAM50 subtype classification, LGG for grade classification in low-grade glioma and KIPAN for kidney cancer type classification. In order to be sure to conduct a fair comparison with MOGONET, we used the same datasets. Wang et al. [12] performed an initial feature selection obtained through the sequential calculation of an ANOVA F-value on the original data to evaluate whether a feature was significantly different across different classes. Moreover, the authors kept the number of features such that the first principal component after feature pre-selection explains at least 50% of the variance.

In the case of ROSMAP and BRCA, as also Gong et al. [13] proceeded, we conducted the analysis on the preprocessed datasets available in Wang et al. [12] GitHub repository. Instead, for LGG and KIPAN, we downloaded the datasets and performed the same preprocessing steps as in Wang et al. [12] since the author did not provide the preprocessed ones. Details on data availability are provided in "Data availability" section. For each of the 5 datasets three types of omics are considered for classification purposes: mRNA expression (mRNA), DNA methylation (meth), and miRNA expression data (miRNA). Table 1 contains all the details for the five datasets.

## Methods

Both in Wang et al. [12] and Gong et al. [13], the authors compared the performance of their methods, MOGONET and MOADNL, respectively, with other typical classification algorithms such as K-nearest neighbours (KNN), Support vector machine (SVM), LASSO regression and block s(PLSDA) as in DIABLO [39].

Taking SVM as an example, the analysis is applied to the concatenation of the 3 multi-omics datasets, where its performance shows a significantly lower accuracy in both studies. However, as SVM can be viewed as a kernel-based classification algorithm, applying it to an early stage integration, i.e., to a combined dataset obtained by simple concatenation of the input datasets, as we have seen, it can be seen as an oversimplification. Moreover, a proper parameters tuning must be carried out along with the choice of a suitable kernel function. Thus, our analysis compares MOGONET's performance with more suitable and fair usage of Multiple kernel learning with support vector machines.

**Table 1** The ROSMAP dataset contains two classes: Alzheimer's disease (AD) patients and normal control (NC). The breast invasive carcinoma dataset (BRCA) contains PAM50 subtype classes: normal-like, basal-like, human epidermal growth factor receptor 2 (HER2)-enriched, Luminal A, and Luminal B. The KIPAN dataset contains different kidney cancer type: chromophobe renal cell carcinoma (KICH), clear renal cell carcinoma (KIRC), and papillary renal cell carcinoma (KIRP). Finally, the LGG dataset is for grade classification in low-grade glioma (LGG)

| Dataset | Classes | Number of features mRNA, meth, miRNA | Features for training mRNA, meth, miRNA |
| --- | --- | --- | --- |
| ROSMAP | NC: 169, AD: 182 | 55,889; 23,788; 309 | 200; 200; 200 |
| BRCA | Normal-like: 115, Basal-like: 131, HER2-enriched: 46, Luminal A: 436, Luminal B: 147 | 20,531; 20,106; 503 | 1000; 1000; 503 |
| KIPAN | KICH: 65; KIRC: 345 ; KIRP: 297 | 60,484; 25,972 ; 1882 | 2000; 2000; 445 |
| LGG | Grade 2: 257 ; Grade 3: 266 | 60,484; 25,972 ; 1882 | 2000; 2000; 548 |

Moreover, new approaches of Multiple kernel learning in combination with deep learning classification models are presented in order to exploit at the same time the adaptability of kernel methods avoiding the optimization of the weights in the convex linear combination and the classification power of deep architectures.

*Multiple kernel learning - SVM*

As presented in "Related work" section, there are many optimization algorithms to compute the coefficients of the convex linear combination of input kernel gram matrices in the literature.

For completeness, in this work, we will present the results obtained using **MKL-naive**, **SimpleMKL** and **SEMKL** in the case of binary classification problem and **STATIS-UMKL**.

On the contrary, STATIS-UMKL in Mariette and Villa-Vialaneix [33] is an algorithm to obtain a consensus meta-kernel in an unsupervised framework. To the best of our knowledge, STATIS-UMKL has never been used with support vector machines for classification purposes. However, the peculiarity of this procedure, which aims to take the different specificities of each dataset into account by fusing them into a single meta-kernel, may also enhance classification performance. In Fig. 1, it is possible to see the network structure for all the SVM algorithms that are used for the experiments. This architecture belongs to the Mixed integration type as the integration of the input omics is preceded by a data transformation, and the SVM algorithm is applied to the convex linear combination of the datasets performed at the feature space.

For completeness, we also trained a support vector machine on the direct concatenation of original datasets (**SVM-concat**) using the same tuning procedure for the hyperparameters used for the other algorithms.

*Deep multiple kernel learning*

As introduced previously, employing neural network architectures is another way to combine the input kernel matrices by avoiding the task of convex linear optimization. More specifically, in Song et al. [14], a deep learning architecture that includes a dense embedding of kernels and a multi-modal neural network is used for fusing multiple kernels.

In our case, we adapted this approach to a multi-omics analysis, meaning that the kernel matrices represent different data sources, i.e different omics, and not different representations of a single data source, as in a classic multiple kernel fusion problem. As shown in Figs. 2 and 3, the structures of the proposed architectures are similar. They consist of a first dense embedding, realized by employing a Kernel PCA for each omic input. After this first step, a multi-modal neural network is used to learn in parallel three representations, one for each dense embedding, and then integrate them to perform the downstream task. In the case of Fig. 2, we call this architecture **Deep Multiple kernel learning**, i.e. **Deep MKL**, to highlight that it is a Multiple kernel learning method that employs deep learning to combine the different kernels information. From a neural network perspective, the architecture is composed of three fully connected layers for each input, followed by an integration step that can be performed through a concatenation, sum, or weighted sum with learnable parameters of the three representations. Finally, another two fully connected layers are employed for the final classification step.

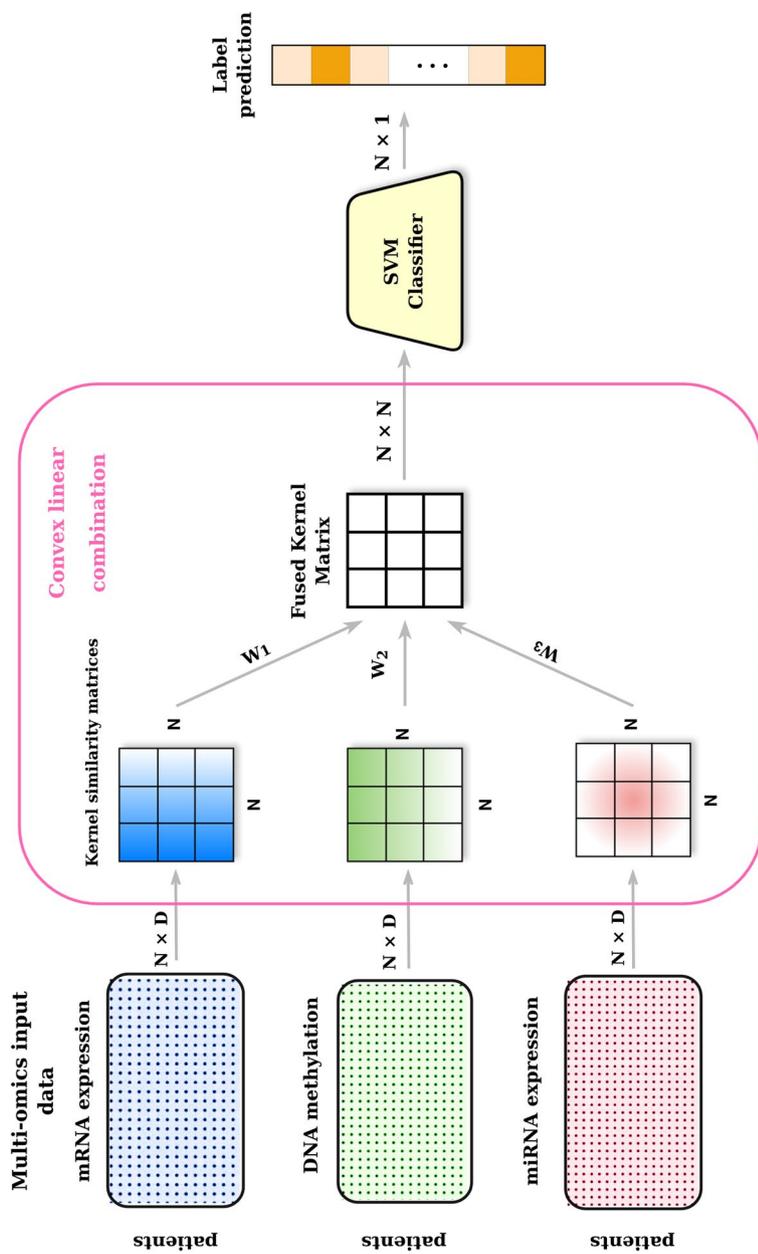

**Fig. 1** A kernel function is applied on each dataset separately. In MKL, a convex linear combination provides a fused Meta-kernel that summarizes the information of input omics. Then an SVM classifier is used for classification

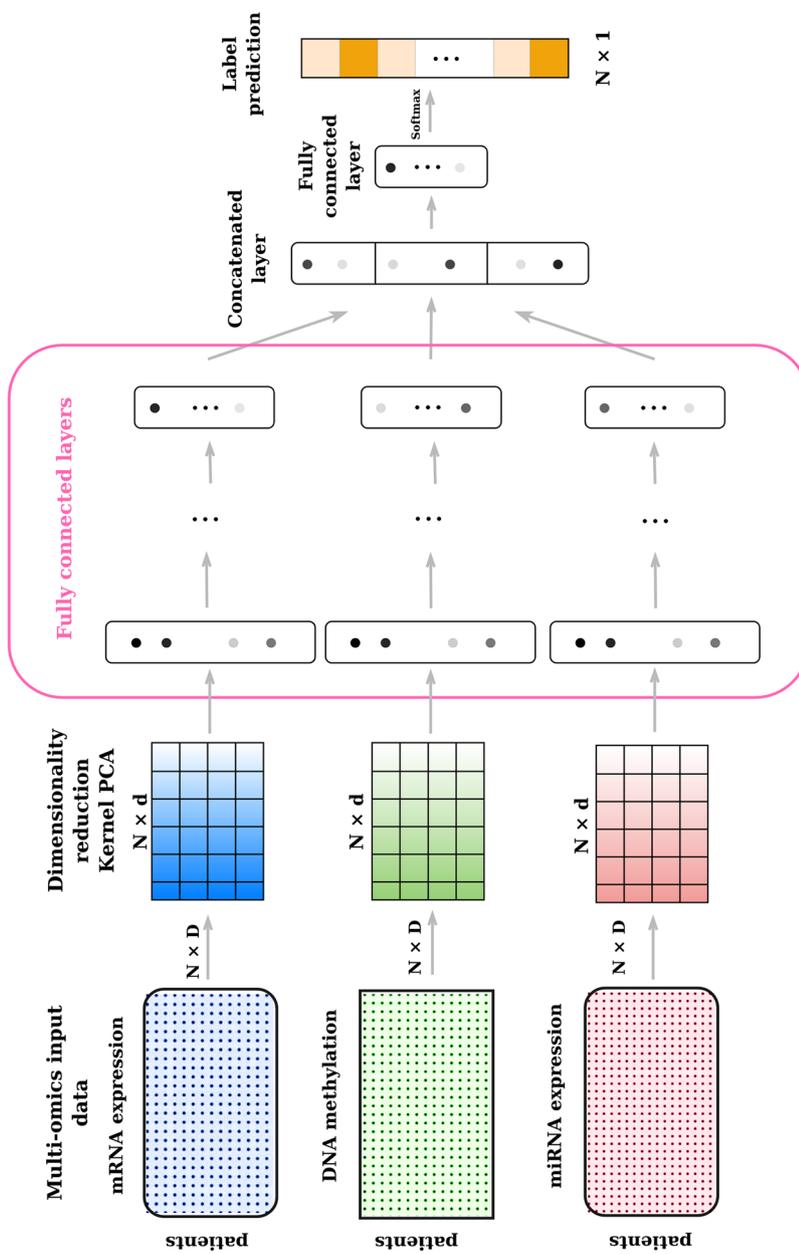

**Fig. 2** Deep MKL (concat) takes in input the Kernel PCA dense embeddings of different omics datasets. It extracts the features using different feedforward sub-networks and then fuses the learnt representations by concatenating them for the final classification

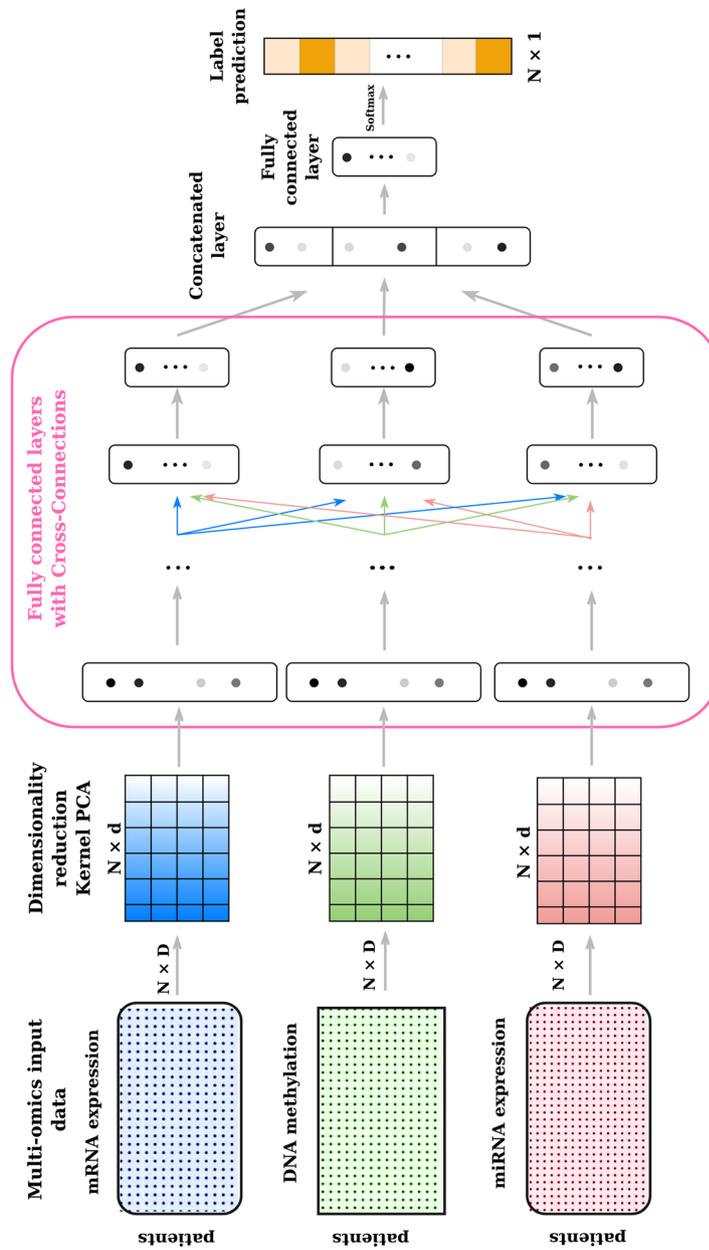

**Fig. 3** Cross-modal Deep MKL (concat) takes in input the Kernel PCA dense embeddings of different omics datasets. It extracts the features using different feedforward sub-networks that are linked by cross-connections, then fuses the learnt representations by concatenating them for the final classification

In the context of multi-modal architectures, cross-connections between modalities can improve the model's performance, allowing the flow of information between modalities at different learning process levels before the fusion step [40, 41]. In our context, this flow should inform each omic layer with each other, potentially improving the performances. We call the version of Deep MKL employing cross-connections **Cross-modal Deep MKL** in Fig. 3. The architecture's structure is similar to the Deep MKL one, except that each cross-connection is, in practice, an additional layer followed by a concatenation step, which means that the Cross-modal Deep MKL architecture, w.r.t. Deep MKL's one, has an additional layer before the integration and classification steps. For both methods, each fully connected layer is followed by a Leaky Relu activation function, a Dropout, and batch normalization. Additional details on the architectures and their specific hyperparameters are discussed in "Hyperparameters tuning" section.

*Interpretability*

Using a dense embedding such as Kernel PCA as a step of a neural network makes the Deep MKL models even more challenging to interpret than classical deep learning ones. In this framework, the principal components can be considered the input features of the neural network. Using an interpretability method such as SHAP in Lundberg and Lee [42] or Integrated Gradients [43] to rank the features would be insufficient because, as highlighted in "Multiple kernel learning" section, after a kernel transformation, the link between the original features, the genes, and the principal components is lost.

In this case, we propose a novel mitigation strategy for biomarkers discovery based on a two-step approach. First, we compute the rank of the input features, namely the kernel principal components, using Integrated Gradients [43] implemented in the library Narine et al. [44]. Then, we employ the recently published method proposed in Briscik et al. [34] to recover the most relevant input variables for the selected principal components. As already introduced in "Multiple kernel learning" section, KPCA-IG in Briscik et al. [34] allows to obtain a data-driven feature ranking based on the selected kPCs, and it is available in the R package *kpcaIG* Briscik et al. [45]. To the best of our knowledge KPCA-IG has never been used in combination with a supervised approach such as our proposed DeepMKL, used to select the most important kernel principal components in terms of prediction accuracy. Combining an unsupervised feature selection approach with a supervised learning method like DeepMKL offers a promising strategy for discovering novel biological and medical biomarkers. This hybrid pipeline may provide deeper insights than traditional methods focused solely on prediction performance, such as those that sequentially remove features to rank their importance based on the impact on prediction accuracy, as in Wang et al. [12] and Gong et al. [13]. We will demonstrate the application of this approach for biomarker identification in "Results" section, highlighting its relevance from a biomedical point of view.

**Performance evaluation**

*Experimental setup*

To evaluate the classification performance of the MKL-SVM algorithms and Deep MKL, we implemented the same evaluation pipeline already used by MOGONET in Wang et al. [12] and by MOADLN in Gong et al. [13]. It consists of evaluating the model's performance on 5 random train/test partitions of the dataset. To maintain the balance of class distributions among the partitions, a stratified version of the split is adopted, keeping the ratio of 30/70 % for the train/test splits.

For final evaluation, we present the mean and standard deviation of different performance metrics among the 5 randomly generated training/test splits, with a seed set of [0, 1, 2, 3, 4] for reproducibility purposes.

The seeds used in MOGONET and Dynamics are not publicly available, meaning that the results are not completely reproducible. For this reason, we recomputed all the metrics using their publicly available code and the same seeds of our experiments in order to have a fair comparison. On the contrary, we have not recomputed the metrics for MOADLN as the code is not publicly available.

*Hyperparameters tuning*

In the context of MKL-SVM, a Grid Search 5-folds cross-validation has been computed on the training sets employing a Gaussian radial basis kernel.

Cross-validation has been used to tune the following parameters:

- *C* parameter: the cost of constraints violation, the so-called C-constant of the regularization term in the Lagrange formulation of the support vector machine algorithm.
- The sigma parameter: the inverse kernel width for the radial basis kernel function.

For the experiments, the *C* parameter has been set in the range [1, 25], while the sigma in the range of [0.005, 0.00005] for both datasets.

In the context of our deep learning methods, we employed a Random Search 5-folds cross-validation for the hyperparameters tuning. Also, in this case, all the experiments were carried out using a Gaussian radial basis kernel for the Kernel PCA step. For all the DeepMKL models, we fixed the number of layers and the number of neurons as in MOGONET, i.e. [200, 200, 100] for ROSMAP and [400, 400, 200] for BRCA, LGG, and KIPAN. For all the Cross-modal Deep MKL architectures, as described in the "Deep multiple kernel learning" section, we implemented cross-connections between modalities, which, in practice, are additional layers. For this reason, we fixed the number of layers and neurons for each dataset as [200, 200, 100, 100] for ROSMAP and [400, 400, 200, 200] for BRCA, LGG, and KIPAN.

In order to have a training process as stable as possible, i.e., a smooth training loss curve, we added a dropout and a batch normalization after each feedforward layer. Additionally, we fixed small values for the learning rate, such as $5 \times 10^{-5}$ for ROSMAP and KIPAN, $10^{-4}$ for BRCA, and $10^{-5}$ for LGG. Regarding the dropout, the intensity is 0.5 for ROSMAP and 0.3 for all the other datasets. Adam classifier [46] and a batch size of 32 are adopted for all the datasets. Regarding the choice of sigma value for the Kernel PCA and the number of principal components to keep, we defined different search spaces for each dataset since the choice of these hyperparameters depends on the topological structure of the data, which varies from dataset to dataset, similar to the *k* parameters used in MOGONET. In the case of ROSMAP, the sigma value for the Kernel PCA is chosen in the set of {0.0005, 0.0007, 0.001}. Meanwhile, for BRCA, the set is {0.00005, 0.0005, 0.005}. For LGG and KIPAN, the set is [0.0005,0.005]. Regarding the number of principal components in ROSMAP, we fixed it to 120. While in BRCA, we defined a search space in the [2, 20] range to choose the optimal combination with the sigma parameter. We adapted the same strategy for LGG and KIPAN using a range of [50, 200].

Since the variability among the different folds made the results unreliable for an early stopping strategy, we chose the number of epochs by defining a range from 100 to 200 with an interval of 10, letting the hyperparameter tuning optimization select the best value in combination with all the other parameters.

For reproducing MOGONET's results, we used the optimized parameter *k*, as suggested by the authors, namely equal to 2 for ROSMAP and 10 for all the other datasets. This parameter controls the average number of edges per node of the Adjacency matrix used for training the graph convolutional neural networks.

Finally, for the DIABLO framework we used the 5-fold cross validation procedure to optimize the number of components (ncomp) for both block PLSDA and block sPLSDA, and the number of retained variables (keepX) for the sparse version. For the design matrix, the value of 0.1 has been used to prioritize the discriminative ability of the model, as suggested by the authors. In Table 2, we provide a summary and description of all the tested methods with all the tuned hyperparameters.

**Table 2** Summary and description of all the tested methods with all the tuned hyperparameters

| Methods | Integration | Optimized Parameters | Description |
| --- | --- | --- | --- |
| block PLSDA | Mixed | ncomp | DIABLO |
| block sPLSDA | Mixed | ncomp, keepX | DIABLO |
| SVM concat | Early | C, $\sigma$ | Direct concatenation |
| SVM naive | Mixed | C, $\sigma$ | Sum of the kernel |
| SimpleMKL-SVM | Mixed | C, $\sigma$ | Weighted sum of kernels |
| SEMKL-SVM | Mixed | C, $\sigma$ | Weighted sum of kernels |
| STATIS-UMKL + SVM | Mixed | C, $\sigma$ | Weighted sum of kernels |
| Deep MKL | Mixed | $\sigma$, epochs, principal components, dropout value | Deep Learning kernel fusion |
| Cross-Modal Deep MKL | Mixed | $\sigma$, epochs, principal components, dropout value | Deep Learning kernel fusion |
| NN_VCDN | Late | NA | Feedforward neural network |
| Dynamics | Late | NA | Dynamical Multimodal Fusion |
| MOGONET | Late | Optimized k | Graph convolutional network |

*Metrics*

In order to have a fair comparison, we employed the same metrics of the state-of-the-art methods. For binary classification, we used accuracy (ACC), F1 score (F1) and area under the curve (AUC):

$$\text{ACC} = \frac{\text{TP} + \text{TN}}{\text{TP} + \text{TN} + \text{FP} + \text{FN}} \qquad (2)$$

with TP = True Positive, TN = True Negative, FP = False Positive and FN = False Negative.

$$\text{F1} = \frac{2 \cdot \text{Precision} \cdot \text{Recall}}{\text{Precision} + \text{Recall}} \qquad (3)$$

where Precision = $\frac{\text{TP}}{\text{TP}+\text{FP}}$ and Recall = $\frac{\text{TP}}{\text{TP}+\text{FN}}$.

In the context of binary classification, the F1 score reflects the harmonic mean between Precision and Recall, meaning that it measures how balanced these other two metrics are for one classifier. The Precision score represents how accurate the positive predictions are. Meanwhile, the Recall metric measures how many True Positives are predicted out of the total number of positive observations.

The AUC score, or area under the ROC curve, represents the classifier's ability to distinguish positive from negative with regard to the classification thresholds. It measures the classifier's performance and its independence from the threshold.

For the multi-class classification task, we employed the accuracy (ACC), the macro-averaged F1 score (F1-macro) and the F1 score weighted by its support i.e. the number of instances in that class (F1-weighted).

In multi-class classification, the F1 score is calculated for each class in a one-vs-all manner. In the case of F1-macro, the F1 scores are then averaged, considering each class equally, regardless of the imbalance of the class distribution in the data.

$$\text{F1-macro} = \frac{1}{C} \sum_{i=1}^{C} F1_i \qquad (4)$$

C is the number of classes and $F1_i$ is the F1 score for the class *i*.

The F1-weighted, instead, takes into account the imbalance of the class distribution in the data, and it is calculated by a weighted average where the weights are the percentage of the instances in one class.

$$\text{F1-weighted} = \frac{1}{C} \sum_{i=1}^{C} \left( \frac{\text{support}_i}{\text{total support}} \right) \cdot F1_i \qquad (5)$$

where $\text{support}_i$ is the number of instances of class *i* and total support is the total number of instances in the data.

## Results

We compared the classification performance of different MKL algorithms with different state-of-the-art methods such as MOGONET and Dynamics, as MOADLN's code is not publicly available. As anticipated, the MOGONET and Dynamics code seeds are unavailable; therefore, we could not replicate the results exactly. Thus, we proceeded with the computation of the metrics for these methods based on the publicly available code and using the same environment and seed selection of our experiments.

Regarding Deep MKL models, we reported the results for only one integration mode, namely *weighted sum*. However, the detailed comparison between different integration modes is provided in the Supplementary Information file.

For BRCA in Table 3, all the MKL algorithms achieved the highest performances for all the metrics. Regarding KIPAN, as shown in Table 6, the MKL algorithms obtained the best results comparable with Dynamics. Also for LGG, the MKL approaches show the best accuracy, where the optimized SVM-concat achieved the best results. On the other hand, for ROSMAP in Table 4, a similar trend can be seen for SVM-based approaches that show comparable accuracy with MOGONET, NN_VCDN and Dynamics, while Deep MKL algorithms perform worse than all the other methods.

Thus, it can be seen that, kernel-based methods are consistently comparable and even outperformed state-of-the-art methods on all four datasets for all the computed performance metrics, Tables 3, 4, 5 and 6.

These results again show the kernel framework's advantages in genomics data mining, where even the results obtained with an SVM trained on the direct concatenation of the input datasets, SVM-concat, exhibits a relatively good performance, especially on ROSMAP and LGG, the smallest datasets. In Wang et al. [12], the performances obtained with SVM-concat are lower, suggesting that even a simple procedure such as early integration followed by proper parameter tuning and an appropriate kernel choice of the SVM may already give a good model alternative for certain datasets. Methods such as SEMKL and STATIS-UMKL, which aim to optimize the input kernel matrices' convex linear combination, showed high performances in most of the different metrics. It should be noted that the MKL with equal weights in SVM-naive showed the best performance in the ROSMAP dataset, indicating that the datasets are probably similarly informative in this context. For this dataset, the second best was STATIS-UMKL + SVM, where the mean over 5 runs of the 3 weights of the convex linear combination of kernel matrices of 0.361, 0.308, 0.331 suggests that the 3 omics are equally important.

**Table 3** Metrics average and standard deviation over 5 random test splits for the performance evaluation on BRCA dataset

|  | BRCA | | |
| --- | --- | --- | --- |
| Algorithm | ACC | F1_weighted | F1_macro |
| block PLSDA | $0.670 \pm 0.016$ | $0.726 \pm 0.009$ | $0.702 \pm 0.011$ |
| block sPLSDA | $0.668 \pm 0.021$ | $0.725 \pm 0.012$ | $0.708 \pm 0.009$ |
| SVM concat | $0.793 \pm 0.018$ | $0.800 \pm 0.016$ | $0.776 \pm 0.017$ |
| SVM naive | $0.838 \pm 0.008$ | $0.849 \pm 0.008$ | $0.828 \pm 0.011$ |
| STATIS-UMKL + SVM | $\mathbf{0.846 \pm 0.011}$ | $\mathbf{0.858 \pm 0.010}$ | $\mathbf{0.837 \pm 0.018}$ |
| Deep MKL (weighted sum) | $0.827 \pm 0.014$ | $0.803 \pm 0.015$ | $0.831 \pm 0.013$ |
| Cross-Modal Deep MKL (weighted sum) | $0.829 \pm 0.017$ | $0.802 \pm 0.022$ | $0.834 \pm 0.015$ |
| NN_VCDN | $0.700 \pm 0.018$ | $0.692 \pm 0.019$ | $0.609 \pm 0.014$ |
| Dynamics | $0.826 \pm 0.010$ | $0.829 \pm 0.010$ | $0.793 \pm 0.020$ |
| MOGONET | $0.736 \pm 0.038$ | $0.726 \pm 0.041$ | $0.650 \pm 0.053$ |

**Table 4** Metrics average and standard deviation over 5 random test splits for the performance evaluation on ROSMAP dataset

| Algorithm | ROSMAP | | |
|---|---|---|---|
| | ACC | AUC | F1 |
| block PLSDA | 0.666 ± 0.025 | 0.689 ± 0.034 | 0.658 ± 0.031 |
| block sPLSDA | 0.671 ± 0.027 | 0.705 ± 0.033 | 0.665 ± 0.017 |
| SVM concat | 0.765 ± 0.019 | 0.863 ± 0.044 | 0.763 ± 0.015 |
| SVM naive | 0.790 ± 0.006 | **0.881 ± 0.010** | 0.778 ± 0.018 |
| SimpleMKL-SVM | 0.758 ± 0.019 | 0.860 ± 0.021 | 0.748 ± 0.012 |
| SEMKL-SVM | 0.775 ± 0.039 | 0.869 ± 0.035 | 0.763 ± 0.037 |
| STATIS-UMKL + SVM | 0.784 ± 0.038 | 0.878 ± 0.019 | 0.772 ± 0.039 |
| Deep MKL (weighted sum) | 0.715 ± 0.028 | 0.800 ± 0.021 | 0.721 ± 0.027 |
| Cross-Modal Deep MKL (weighted sum) | 0.730 ± 0.025 | 0.802 ± 0.020 | 0.746 ± 0.039 |
| NN_VCDN | **0.794 ± 0.030** | 0.874 ± 0.024 | 0.807 ± 0.036 |
| Dynamics | 0.764 ± 0.026 | 0.870 ± 0.011 | 0.771 ± 0.031 |
| MOGONET | 0.787 ± 0.027 | 0.878 ± 0.021 | **0.791 ± 0.045** |

**Table 5** Metrics average and standard deviation over 5 random test splits for the performance evaluation on LGG dataset

| Algorithm | LGG | | |
|---|---|---|---|
| | ACC | AUC | F1 |
| block PLSDA | 0.651 ± 0.024 | 0.713 ± 0.034 | 0.677 ± 0.029 |
| block sPLSDA | 0.637 ± 0.030 | 0.771 ± 0.039 | 0.692 ± 0.027 |
| SVM concat | **0.723 ± 0.030** | **0.781 ± 0.024** | **0.741 ± 0.032** |
| SVM naive | 0.709 ± 0.011 | 0.774 ± 0.024 | 0.724 ± 0.022 |
| SimpleMKL-SVM | 0.684 ± 0.011 | 0.759 ± 0.024 | 0.710 ± 0.020 |
| SEMKL-SVM | 0.691 ± 0.011 | 0.762 ± 0.028 | 0.719 ± 0.017 |
| STATIS-UMKL + SVM | 0.709 ± 0.009 | 0.774 ± 0.023 | 0.728 ± 0.015 |
| Deep MKL (weighted sum) | 0.687 ± 0.011 | 0.765 ± 0.025 | 0.684 ± 0.031 |
| Cross-Modal Deep MKL (weighted sum) | 0.700 ± 0.020 | 0.768 ± 0.026 | 0.695 ± 0.032 |
| NN_VCDN | 0.703 ± 0.036 | 0.754 ± 0.030 | 0.715 ± 0.028 |
| Dynamics | 0.707 ± 0.029 | 0.769 ± 0.027 | 0.714 ± 0.023 |
| MOGONET | 0.669 ± 0.026 | 0.711 ± 0.026 | 0.69 ± 0.032 |

As expected, the two wrapper methods optimized for supervised multiple kernel learningn namely, SimpleMKL and SEMKL seem to have similar performance as already shown in Wilson et al. [15]. On the other DIABLO linear approaches showed lower performances, proving the need of nonlinear based approaches in the context of complex omics datasets. The Deep MKL approach to integrating multiple kernels shows results comparable with the STATIS-UMKL + SVM method for the BRCA, LGG, and KIPAN datasets. In the case of the ROSMAP dataset, it performs worse than all the methods based on SVM. The difference in performance can be largely attributed to the dataset sizes. This phenomenon is consistent with established understanding that deep learning models tend to underperform in scenarios involving smaller datasets [47].

**Table 6** Metrics average and standard deviation over 5 random test splits for the performance evaluation on KIPAN dataset

| Algorithm | KIPAN | | |
| --- | --- | --- | --- |
| | ACC | F1_weighted | F1_macro |
| block PLSDA | 0.882 ± 0.013 | 0.884 ± 0.013 | 0.871 ± 0.016 |
| block sPLSDA | 0.896 ± 0.012 | 0.898 ± 0.011 | 0.891 ± 0.017 |
| SVM concat | 0.953 ± 0.010 | 0.954 ± 0.009 | 0.949 ± 0.020 |
| SVM naive | 0.958 ± 0.010 | 0.959 ± 0.009 | 0.953 ± 0.018 |
| STATIS-UMKL + SVM | 0.959 ± 0.010 | **0.960 ± 0.010** | 0.955 ± 0.017 |
| Deep MKL (weighted sum) | 0.958 ± 0.011 | 0.954 ± 0.018 | 0.958 ± 0.011 |
| Cross-Modal Deep MKL (weighted sum) | 0.958 ± 0.009 | 0.952 ± 0.014 | **0.958 ± 0.009** |
| NN_VCDN | 0.957 ± 0.006 | 0.957 ± 0.006 | 0.952 ± 0.015 |
| Dynamics | **0.960 ± 0.011** | 0.960 ± 0.010 | 0.951 ± 0.022 |
| MOGONET | 0.940 ± 0.023 | 0.932 ± 0.032 | 0.941 ± 0.023 |

Cross-connections, which were expected to improve the predictions as they ensure more layers of integration between different omics, show no consistent improvement w.r.t. the simpler Deep MKL architecture.

**Biomarker discovery**

We previously introduced the approach for biomarkers discovery employing a hybrid 2-step approach for the Deep MKL algorithm. First, the most relevant features, i.e., kernel principal components, are selected using Integrated Gradients [43] and subsequently KPCA-IG as in Briscik et al. [34] is applied, obtaining a data-driven feature importance based on the kernel PCA representation of the data. The optimal tuned $\sigma$ parameters adopted in the Deep MKL model are also used to run the KPCA-IG method. The most important biomarkers can be found in Tables 7 and 8.

For BRCA dataset the most important components are [1, 2, 3], [2, 1, 3] and [2, 1, 4] for mRNA, meth and miRNA respectively. As the mRNA influence on the final prediction appeared to be more prominent, we included the first 15 most relevant genes, while we showed the first 10 for the DNA methylation and miRNA datasets. Same procedure is applied to the ROSMAP dataset where the most relevant components are [1, 2, 21], [1, 2, 3] and [1, 4, 9] for the three datasets respectively. For the mRNA expression genes and those inferred from high-ranking DNA methylation features, we conducted gene set functional enrichment analysis using the ToppGene Suite [48] to assess the biological significance of genes identified by Deep MKL, highlighting biological annotations such as Gene Ontology (GO) terms that are significantly enriched in a specific set of genes. To correct for multiple comparisons and control the false discovery rate (FDR), the Benjamini-Hochberg procedure is employed, reporting the adjusted p-values.

**Table 7** Important biomarkers identified by DeepMKL + KPCA-IG in the BRCA dataset

| Omics data type | Biomarkers |
| --- | --- |
| mRNA expression (15) | GABRP, SOX10, TFF1, KRT6B, AGR3, KLK7, SERPINB5, DSC3, KLK6, AGR2, MIA, TRIM29, SLC6A14, KRT16, KLK8 |
| DNA methylation (10) | IGFBP4, RARA, NHLRC4, CA12, DNALI1, MIR26B, GPR37L1, RSAD1, RARG, NR2F6 |
| miRNA expression (10) | hsa-mir-224, hsa-mir-452, hsa-mir-505, hsa-mir-675, hsa-mir-577, hsa-mir-375, hsa-mir-18a, hsa-mir-196b, hsa-mir-511-2, hsa-mir-145 |

**Table 8** Important biomarkers identified by DeepMKL + KPCA-IG in the ROSMAP dataset

| Omics data type | Biomarkers |
| --- | --- |
| mRNA expression (15) | PREX1, CSRP1, MID1IP1, PLXNB1, MINDY1, SLC44A1, ANLN, CAVIN1, SLC6A9, DOCK5, ITPKB, SASH1, YES1, CLMN, CARHSP1 |
| DNA methylation (10) | R3HDML, MYOD1, HYAL2, ALDH3B1, OTOP3, CHST14, GPR152, LAG3, ENG, MYO1C |
| miRNA expression (10) | hsa-miR-423-3p, hsa-mir-374b, hsa-miR-487b, hsa-miR-361-5p, hsa-miR-30b, hsa-miR-885-5p, hsa-miR-376a, hsa-miR-216a, hsa-miR-548b-3p, hsa-miR-26a |

For BRCA PAM50 subtype classification datasets, several of the 15 selected genes from the mRNA expression dataset were included in GO terms linked with breast cancer such as $\beta$-alanine transmembrane transporter activity (GO:0001761, $p = 2.324E-2$), carnitine transmembrane transporter activity (GO:0015226, $p = 4.953E-2$) and dystroglycan binding (GO:0002162, $p = 3.759E-3$). For instance, $\beta$-alanine has been targeted for its several anti-tumor effects and as a co-therapeutic agent in the treatment of breast tumors [49]. Moreover, the gene SLC6A14 involved in the $\beta$-alanine and carnitine transmembrane transporter activities has already been addressed to have a pivotal role in the cancer stage [50], where its deletion has been linked to a reduction of cancer growth and metastatic spread [51], thus being selected as potential direct drug target for cancer therapy [52]. Also, the dystroglycan binding has been linked with breast cancer as the expression of this adhesion molecule is frequently reduced in human breast and colon cancers and is associated with tumor progression [53]. Within this GO, the two enriched genes that we found are AGR3 and AGR2. For instance, AGR3 had already been characterized as a novel potential biomarker both for breast cancer prognosis and early breast cancer detection [54], while AGR2 expression has been correlated with poor outcomes of patients with ER-positive breast cancer [55]. Among others, SERPINB5, DSC3, and GABRP have also been linked with malignant neoplasms of the breast. SERPINB5 has been indicated to inhibit tumor progression [56], DSC3 downregulation has been linked with several cancer types [57] and GABRP over-expression has been linked with poor prognosis, metastatic cancer, basal-like breast cancer [58–60]. For genes related to the identified DNA methylation features, several interesting GO were enriched, including prosaposin receptor activity (GO:0036505, $p = 1.607E-2$) and insulin-like growth factor II binding (IGF-2) (GO:0031995, $p = 4.011E-2$). Several studies have shown that prosaposin, a regulator of estrogen receptor alpha, promotes breast cancer growth [61, 62] and that IGFs play an important role in cancer development [63] and specifically and increased IGF-2 production has been linked with cancer development and progression in many conditions [64–67]. Moreover, the highly-ranked miRNAs selected by our method have also exhibited an association with cancer. Zhang et al. [68] found over-expression of hsa-miR-224 in breast cancer cell lines and in TNBC primary cancer samples. Another example is hsa-mir-675 as in Vennin et al. [69] it has been shown that over-expression of this miRNA enhances the aggressive phenotype of breast cancer cells, including increased cell proliferation and migration in vitro and increased tumor growth and metastasis in vivo.

Deep MKL with KPCA-IG also identifies important biomarkers related to Alzheimer's disease. For AD patient classification, for genes identified by mRNA expression features, several enriched GO has been linked with the Alzheimer patalogy. For instance inositol-1,4,5-trisphosphate 3-kinase activity (GO:0008440, $p = 3.912E-2$) linked with the gene ITPKB, it has been found to increase in human Alzheimer brain and to exacerbates mouse Alzheimer pathology [70]. Also the choline transmembrane transporter activity (GO:0015220, $p = 4.110E-2$) as been showed to be linked with the disease, as the choline transporter was marked to be incremented in cortical brain regions from AD patients compared to non-AD control [71], as also the gene involved in the signature, namely SLC44A1 has been found to be up-regulated in Alzheimer patients [72]. Moreover, the first gene in the list, namely PREX1 has been reported to be linked with brain-related conditions, such as aberrant neuronal polarity and psychosis-related behaviors, in case of over-expression [73].

Additionally, the GO transforming growth factor beta binding (GO:0050431, $p = 2.69E-2$) was enriched for genes linked to the selected DNA methylation features by our procedure. Dysfunction in TGF$\beta$ signaling has been linked to exacerbated neuroinflammation promoting microglia's cytotoxic activation, which may contribute to neurodegeneration in AD [74]. Moreover, several genes are significantly annotated in aldehyde dehydrogenase (NADP+) activity (GO:0033721, $p = 2.911E-2$) where aldehyde dehydrogenase two activity and aldehydic load has been associated to a contribution in neuroinflammation and Alzheimer's disease-related pathology [75]. Another molecular function is the protein tyrosine kinase inhibitor activity (GO:0030292, $p = 3.271E-2$). It has been shown that tyrosine kinase inhibition can be viewed as a potential target for therapeutic intervention for treating Alzheimer's disease as it represents a valid mechanism for improving autophagic clearance of neurotoxic protein and mitigating mast cell and microglial-mediated inflammation [76]. Other GO potentially related to AD are hexosaminidase activity (GO:0015929, $p = 4.290E-2$) and galactose binding (GO:0005534, $p = 3.271E-2$), where abnormal cortical lysosomal $\beta$-hexosaminidase and $\beta$-galactosidase activity has been linked both to early and the advanced stage of Alzheimer's disease [77]. Regarding the miRNA biomarkers, our methods selected, among others, hsa-miR-361-5p, which was found to be abnormally expressed in AD patients [78]. Another highly-ranked miRNA, hsa-miR-885-5p, is substantially expressed in brain tissues and has been associated with AD [79].

## Conclusion

Multiple kernel learning is a well-established algorithm in the machine learning community, while its use has yet to be widespread among practitioners for bio-data mining.

This work presents two novel different approaches for Multiple kernel learning in the context of multi-omics data integration. One employs unsupervised learning techniques along with Support Vector Machines (SVM). The other utilizes deep learning as a substitute for convex linear optimization to integrate kernels. The proposed methodologies are tested and compared with state-of-the-art methods performances. The experimental results on four publicly available biomedical datasets show that approaches based on kernel mixed integration exhibit comparable or even improved performance while being considerably simpler. Precisely, we demonstrate that MKL methods show competitive results compared to claimed state-of-the-art methods, which failed to improve predictive performance in the context of multi-omics analysis. Also the novel deep learning-based procedures used to integrate input kernels and for classification demonstrate to be a valid alternative to the more classical Multiple kernel learning optimizations in the case of datasets with large enough sample size. In addition, we proposed a novel method for biomarkers discovery based on our newly proposed Deep MKL method, which proved effective for predicting the disease of interest, potentially showing disease mechanisms and helping in the development of personalized treatment protocols.

Future work could investigate other types of data kernel embedding and different deep architectures to exploit the kernel framework in the context of Deep multiple kernel learning. For classical multiple kernel learning, different types of kernel functions can be tested, as each omic dataset could benefit from ad-hoc kernel function choices.

MKL showed that despite being under-utilized in multi-omics data analysis, it provides a fast and reliable solution that can compete with and outperform more complex architectures.

## Supplementary Information

The online version contains supplementary material available at https://doi.org/10.1186/s13040-024-00406-9.

Supplementary Material 1.

**Acknowledgements**
Not applicable.

**Authors' contributions**
MB and GT wrote the main manuscript and performed the analysis under the supervision of SD, MAD and LV. All authors read and approved the final manuscript.

**Funding**
This work was funded by E-MUSE MSCA-ITN-2020 European Training Network under the Marie Sklodowska-Curie grant agreement No 956126. The funding did not influence the study's design, the interpretation of data, or the writing of the manuscript.

**Data availability**
The LGG and KIPAN datasets were obtained from UCSC Xena Goldman et al. [80] https://xenabrowser.net/datapages/.
 All the reduced datasets are available in the GitHub repository https://github.com/txWang/MOGONET/tree/main.
 The source code of this work can be downloaded from GitHub: https://github.com/gabrieletaz/MKL_MO.

## Declarations

**Ethics approval and consent to participate**
Not applicable.

## Publisher's Note